\documentclass[a4paper,fleqn]{cas-dc}

\usepackage[authoryear,longnamesfirst]{natbib}
\usepackage{amsmath}
\usepackage{amsthm}
\usepackage{amssymb}
\usepackage{amsfonts}
\usepackage{booktabs}
\usepackage{subcaption}
\usepackage{url}
\usepackage{siunitx}
\usepackage{graphicx}
\usepackage{xcolor}

\theoremstyle{plain}
\newtheorem{theorem}{Theorem}[section]

\theoremstyle{definition}
\newtheorem{definition}[theorem]{Definition}

\theoremstyle{remark}

\graphicspath{{figures/}{./}}

\begin{document}
\let\WriteBookmarks\relax
\def\floatpagepagefraction{1}
\def\textpagefraction{.001}

% Short title (running head)
\shorttitle{STAR-XPC: Set Transformer for X-ray Transmission in Polymer Composites}

% Short author (running head)
\shortauthors{C. Erer et al.}

\title [mode = title]{STAR-XPC: Set Transformer for Estimating X-ray Transmission in Polymer Composites}

\author[1]{Can Erer}
\ead{salih.erer@std.bogazici.edu.tr}

\affiliation[1]{organization={Department of Computer Engineering, Boğaziçi University},
            addressline={Bebek},
            city={Istanbul},
            postcode={34342},
            country={Turkiye}}

% Second author (corresponding)
\author[1]{H. Birkan Yilmaz}
\cormark[1]
\ead{birkan.yilmaz@bogazici.edu.tr}

\author[2]{Nebahat Aral}
\ead{nebahataral@gmail.com}

\affiliation[2]{organization={Independent Researcher},
            city={Istanbul},
            country={Turkiye}}

\cortext[1]{Corresponding author}

% Abstract
\begin{abstract}
Evaluating the energy transmission ratio of an X-ray beam through a polymer composite is a challenging task in radiation shielding studies. The reference solution with Monte Carlo transport simulation is highly accurate but computationally expensive: each of our reference simulations is capped at $4\times 10^{9}$ photon histories and a $20\,000$~s (over $5$ hours) time limit, while closed-form analytical methods fail to represent such complex scenarios. We present an artificial neural network model that maps the unordered multiset of molecular constituents, their mass fractions, and four bulk physical descriptors directly to the transmission ratio $I(t)/I_{0}$ in 1-3  milliseconds. The model is a Set Transformer, which makes it permutation invariant by construction and accepts mixtures of different sizes up to 11 constituents within a fixed-size padded set. Each constituent is embedded with a chemistry-pretrained transformer language model, ChemBERTa. In our proposed method, training data are produced by an automated PENELOPE/penEasy pipeline using the radiation source at $30$, $80$, $100$, and \SI{120}{\kilo\volt} tube voltages over polyethylene-based shielding composite slabs of thicknesses spanning $0.15$-$2.5$~mm. After preprocessing, the dataset contains $n=4697$ simulated configurations on which a leakage-free, jointly stratified 10-fold cross-validation is performed, attaining mean squared error $\mathrm{MSE} = (1.12\pm 0.56)\times 10^{-4}$, normalized-MSE percentage $\mathrm{NMSE\%}=0.0294\%\pm 0.0147\%$, and $R^{2}=0.9988$. The corresponding per-sample percentage errors give an $\alpha$-trimmed ($\alpha=0.10$) absolute percentage error (APE) of $1.66\%$ and a median APE of $1.15\%$. Our model's inference is several orders of magnitude faster than the underlying Monte Carlo simulation (sub-millisecond per query  vs.\ several hours for the reference run) with effective accuracy in predicting the $I(t)/I_{0}$.
\end{abstract}

% Keywords (separated by \sep)

\maketitle

	%%%%%%%%%%%%%%%%%%%%%%%%%%%%%%%%%%%%%%%%%%%%%%%%%%
    %%%%%%%%%%%%%%%%%%%%%%%%%%%%%%%%%%%%%%%%%%%%%%%%%%
    %%%%%%%%%%%%%%%%%%%%%%%%%%%%%%%%%%%%%%%%%%%%%%%%%%
    \section{Introduction}

    Calculating how a solid reduces the intensity of an X-ray beam is one of the most crucial tasks for those who work in diagnostic imaging, image-guided radiotherapy, industrial radiography, and the engineering of radiation shields in medical or nuclear fields. In practical applications, material designers rarely work with a single material, indeed they mostly work with composites, and polymer-matrix composites loaded with high-$Z$ fillers such as Bi$_{2}$O$_{3}$, WO$_{3}$, or BaSO$_{4}$ are an active area of radiation-shielding research in their own right~\citep{more2021polymer,verma2023review,aral2016alternative,aral2017x}.

    There are three characteristics that actually make this task complex and non-trivial. First, the composites consist of different constituents, which we call being \emph{heterogeneous}: a thin slab contains molecules with interaction cross sections that differ by orders of magnitude across the kilovoltage range. Secondly, the source emits many wavelengths at once, which we call being \emph{polychromatic}: when we choose an energy level between 30-\SI{120}{\kilo\volt} the distribution belonging to that energy level varies a lot. Lastly, there are various choices for possible designs, which we call being \emph{combinatorial}: a mix of materials may consist of arbitrary materials and may have arbitrary thickness levels. Due to the vast design space, engineers must test thousands of candidates in an optimization loop. Hence, engineers need accurate and reliable methods to know how much energy passes through the given material based on its physical properties such as its shape and constituents.

    For heterogeneous mixtures and polychromatic energy spectrum, there is no closed-form analytical solution. Mass attenuation mixture rules evaluated with tabulated data, such as NIST's X-ray mass attenuation tables (SRD 126, a.k.a. XAAMDI)~\citep{hubbell2004xcom} or cross-section tools such as XCOM~\citep{berger1987xcom} and its derivatives WinXCom~\citep{gerward2001winxcom} and Phy-X/PSD~\citep{sakar2020phyx}, are fast but inherit two restrictive assumptions: narrow-beam geometry and a monoenergetic source. In broad-beam geometry, scattered photons reaching the detector must be accounted for with empirical buildup factors~\citep{harima1993buildup}, which are tabulated for specific materials and energies and are not readily available for arbitrary multi-component composites. XAAMDI tabulates mass attenuation coefficients by mass-weighting constituent elements at a single photon energy, which works only if the material is treated as one uniform blend. Moreover, it gives one coefficient per energy, not a spectrum response, so using it for a 30-120 kV polychromatic beam means manually querying and integrating many energy points, with no account for how the spectrum hardens as it penetrates the material.

    PENELOPE/penEasy~\citep{salvat2019penelope,sempau2011penelope}  uses Monte Carlo methods in order to simulate the interaction utilizing the physical laws. The estimations of those codes are accurate, they are used as a reference to illustrate how radiation attenuates throughout the given material with respect to its physical properties. The data they have and the algorithms used are extensively tested and concluded to be reliable. However, their cost is prohibitive for use in an optimization loop. Each reference run is configured with a nominal cap of $4\times 10^{9}$ photon histories, a $2\,\%$ relative-uncertainty target, and a $20\,000$~s time limit. In practice, the time limit is the binding constraint for most configurations, so the number of histories actually completed, and the resulting precision, vary by configuration rather than uniformly reaching the nominal targets. The statistical error decays only as $\sigma\propto M^{-1/2}$ in the number of histories. Obtaining a reliable estimation  requires a lot of computational power, hence it is not possible to use these codes to simulate all combinations in our design space.

    In short, current limitations mainly stem from the computational cost of running Monte Carlo simulations and the fact that we do not have an analytical solution for direct calculation. The capabilities of classical machine learning (ML) methods can be considered as a secondary limitation.

    The attenuation of a homogeneous mixture does not depend on the order of its constituents, so the target is a permutation-invariant function of the constituent set. Multi-layer perceptrons over concatenated slots and sequence-based models impose an artificial ordering and must learn this invariance from data, spending parameters and samples on repeating the same mapping for every ordering. Tree-based regressors share this problem once the constituents are flattened into fixed columns, and they treat each embedding dimension as an independent feature, which suits dense embeddings poorly. A canonical ordering, e.g.\ by mass fraction, restores invariance but is arbitrary and changes abruptly when two fractions cross. We therefore build the invariance into the architecture.

	This paper proposes a permutation-invariant set transformer for estimating X-ray transmission in polymer composites, namely STAR-XPC. Our proposed method treats a composite as an unordered set of tokens enriched with a vector of bulk physical descriptors, and predicts the energy transmission ratio. The contributions of this work are:
	\begin{enumerate}
        \item An automated PENELOPE/penEasy dataset-generation pipeline \footnote{\url{https://github.com/cmpe-netlab41/attenupy}}.
        \item A chemistry-aware embedding pipeline that summarizes each constituent with the attention-mask-weighted mean of the last four hidden layers of ChemBERTa~\citep{chithrananda2020chemberta} (a 768-dimensional vector), followed by $\ell_{2}$ normalization. Aggregating the upper layers captures both syntactic and semantic features of the simplified molecular input line entry system (SMILES) string used in cheminformatics.
		\item A permutation-invariant Set Transformer model based on Induced Set Attention Blocks (ISAB) and Pooling by Multihead Attention (PMA), which predicts the energy transmission ratio in 1-3 milliseconds. The STAR-XPC model and inference codes are available at HuggingFace\footnote{ \url{https://huggingface.co/cmpe-netlab41/STAR-XPC}}.
	\end{enumerate}

    %%%%%%%%%%%%%%%%%%%%%%%%%%%%%%%%%%%%%%%%%%%%%%%%%%
    %%%%%%%%%%%%%%%%%%%%%%%%%%%%%%%%%%%%%%%%%%%%%%%%%%
    %%%%%%%%%%%%%%%%%%%%%%%%%%%%%%%%%%%%%%%%%%%%%%%%%%
	\section{Background Information}\label{sec:theory}
	\subsection{Radiation Attenuation Physics}

	For a narrow, monoenergetic photon beam of energy $E$ and intensity $I_{0}(E)$ traversing a homogeneous slab of thickness $t$, Beer-Lambert law states that
	\begin{equation}
		I(t,E) = I_{0}(E)\,e^{-\mu(E)\,t},
		\label{eq:bl}
	\end{equation}
	where $I(t,E)$ is the transmitted intensity for the input thickness and energy, $\mu(E)$ is the linear attenuation coefficient for the input energy~\citep{attix1986radiological}. For a multi-component mixture, the standard mass-additivity rule for the mass attenuation coefficient is
	\begin{equation}
		\bigl(\mu/\rho\bigr)_{\mathrm{tot}}(E) = \sum_{i=1}^{N} w_{i}\,\bigl(\mu/\rho\bigr)_{i}(E),
		\qquad \sum_{i=1}^{N} w_{i} = 1,
		\label{eq:mixrule}
	\end{equation}
	where $w_{i}$ is the mass fraction of constituent $i$. For a polychromatic source spectrum $S\!(E)$ incident on a slab of bulk density $\rho$ and thickness $t$, the integrated energy transmission ratio is
	\begin{equation}
		 y(t)\;\equiv\;\frac{I(t)}{I_{0}}
		\;=\;\frac{\displaystyle\int_{0}^{E_{\max}} E\,S\!(E)\,e^{-(\mu/\rho)_{\mathrm{tot}}(E)\,\rho\,t}\,dE}
		{\displaystyle\int_{0}^{E_{\max}} E\,S\!(E)\,dE}
		\label{eq:rar}
	\end{equation}
	where $I(t)$ is the transmitted intensity through a slab of thickness $t$ and $I_{0}$ is the incident intensity. Equation~\eqref{eq:rar} gives the narrow-beam idealization, but our target is its Monte Carlo counterpart, which additionally captures scattered photons and characteristic X-ray fluorescence reaching the scoring region. This is the target variable for our model, i.e., our model is trained to predict the energy transmission ratio $y(t)$.

	In the kilovoltage range, X-ray attenuation is governed by three interaction channels: photoelectric absorption ($\tau \propto Z^{n}/E^{m}$, $n\in[4,5]$ and $m\in[3,3.5]$), incoherent (Compton) scattering, and coherent scattering~\citep{attix1986radiological}. At a fixed energy, the mass-additivity rule (Eq.~\eqref{eq:mixrule}) is exact under the standard assumption that the atomic wavefunctions in the mixture are unaffected by the surrounding chemical or crystalline environment. Combined with Beer-Lambert's law under narrow-beam (for a monoenergetic beam in narrow-beam/scatter-free) geometry, this gives an exact analytic prediction of transmission. However, two effects break this shortcut in practice: beam hardening reshapes the spectrum $S\!(E)$ along the depth coordinate as low-energy photons are preferentially absorbed, and near an element's K-edge, the mass attenuation coefficient jumps sharply over a narrow energy window, so any closed-form spectral integration must resolve this discontinuity precisely; moreover, the characteristic fluorescence photons emitted following K-shell photoelectric absorption are transported and re-absorbed within the surrounding matrix in a way no closed-form correction captures. These effects are jointly resolved via the Monte Carlo simulation method.

	\subsection{Monte Carlo Simulation Method}\label{sec:penelope}

	PENELOPE~\citep{salvat2019penelope} (PENetration and Energy LOss of Positrons and Electrons) is a Fortran subroutine library that performs coupled photon-electron-positron transport in arbitrary materials and quadric geometries via a \emph{mixed simulation} algorithm. Hard interactions (those with polar deflection or energy loss exceeding pre-selected cutoff values) are sampled individually, while soft interactions are condensed into a continuous-slowing-down approximation governed by the user-controlled cutoff parameters. PenEasy~\citep{sempau2011penelope} is a structured general-purpose main program that drives PENELOPE through a plain-text input file and exposes a set of pre-implemented source models, geometry handlers, tallies, and variance reduction techniques.

	For a stationary and time-independent source, the photon angular flux $\Phi(\mathbf{r},E,\boldsymbol{\Omega})$ satisfies the linear Boltzmann transport equation
	\begin{equation}
		\begin{split}
			\boldsymbol{\Omega}\cdot\nabla\Phi
			+ \Sigma_{t}(\mathbf{r},E)\,\Phi
			&= \iint \Sigma_{s}(E'\!\to\!E,\,\boldsymbol{\Omega}'\!\to\!\boldsymbol{\Omega})\\
			&\qquad\times\Phi(\mathbf{r},E',\boldsymbol{\Omega}')\,dE'\,d\boldsymbol{\Omega}'\\
			&\qquad+ Q(\mathbf{r},E,\boldsymbol{\Omega}),
		\end{split}
		\label{eq:btransport}
	\end{equation}
	where $\Sigma_{t}$ is the macroscopic total cross-section, $\Sigma_{s}$ the
	differential scattering kernel, and $Q$ the volumetric source term.
	PENELOPE samples particle histories from Eq.~\eqref{eq:btransport} via
	the standard Markov-process construction.

    Each history costs a fixed average time, so ${T \!\propto\! M}$, while the statistical uncertainty of an estimator built from $M$ independent histories decays as $\sigma_{\hat\theta} \propto M^{-1/2}$. Eliminating $M$ yields,
    %%%%%%%%%%%%%%%%%%
    \begin{equation}
    T \;\propto\; \sigma_{\hat\theta}^{-2},
    \label{eq:cost}
    \end{equation}
    so halving the uncertainty quadruples the cost and, conversely, a $10\times$ speed-up can only be bought at a $3.2\times$ larger uncertainty. Our runs use up to $4\!\times\!10^{9}$ histories and take approximately $\sim$5 CPU-hours, with almost all of them reaching a relative uncertainty of 2\%. No reduction of $M$ brings a converged transport run into the range required by an inner design-optimisation loop, which is what motivates a learned ML model.

	\subsection{Machine Learning for Physics Applications}\label{sec:mlphys}

    ML modeling of costly simulators is a standard approach in computational physics. Successful examples include deep operator networks that learn PDE solution operators directly from data~\citep{lu2021deeponet,li2021fno}. The MC radiation transport literature is dominated by tabular regressors, with occasional deep-learning exceptions in adjacent domains such as CT dose reconstruction~\citep{salimi2023realtime}. Within radiation shielding specifically, ANNs have been trained to predict mass or linear attenuation coefficients of glass systems~\citep{benhadjira2024ann} and of Bi$_2$O$_3$-based and multi-filler composites from Geant4-generated data~\citep{alipoor2026bi2o3,alipoor2026aprons}. These models take a fixed-length feature vector (e.g.\ photon energy and density) and predict a monoenergetic coefficient, whereas we predict the spectrum-integrated transmission of a polychromatic tube beam directly from an unordered set of constituents. None of these architectural preferences systematically encode the symmetry of a molecular mixture, the ordering of constituents in the input has no physical meaning. The statistical error of an unbiased Monte Carlo estimator calculated using M histories decays at the standard $M^{-1/2}$ rate, which makes high-precision MC essentially costly and motivates the utilization of an ML model (ML surrogate).

    An aggregate property of an \emph{unordered multiset} of molecular constituents is what we are trying to estimate. Deep learning on flattened tabular vectors does not fit this structure, since assigning constituents to fixed column positions makes the prediction depend on an arbitrary slot order, and recurrent networks would likewise introduce an ordering. Padding a set to a fixed size is compatible with permutation invariance only if the padding tokens are interchangeable and the model is itself permutation invariant, as in our architecture (Section~\ref{sec:arch}).

	\begin{definition}[Permutation invariance]
		A function $f$ acting on a finite multiset $\mathcal{S}=\{\mathbf{x}_{1},\ldots,\mathbf{x}_{n}\}\subset\mathbb{R}^{d}$ is \emph{permutation invariant} if, for every permutation $\pi\in\mathfrak{S}_{n}$,
		\begin{equation*}
			f\!\bigl(\{\mathbf{x}_{1},\ldots,\mathbf{x}_{n}\}\bigr) = f\!\bigl(\{\mathbf{x}_{\pi(1)},\ldots,\mathbf{x}_{\pi(n)}\}\bigr).
		\end{equation*}
	\end{definition}

	While the Deep Sets~\citep{zaheer2017deepsets} guarantees universality, its sum-decomposition requires a latent dimension at least as large as the set cardinality, which can make the simple sum aggregator impractical for larger sets~\citep{wagstaff2019limitations}. The Set Transformer~\citep{lee2019settransformer} replaces the sum aggregator with attention.

	We used Multi-Head Attention (MHA)~\citep{vaswani2017attention} and Induced Set Attention (ISA) blocks~\citep{lee2019settransformer} in our ML surrogate. Let $\mathbf{Q}\in\mathbb{R}^{n_{q}\times d}$, $\mathbf{K},\mathbf{V}\in\mathbb{R}^{n_{k}\times d}$ be query and key-value matrices. Let $h$ be the number of heads and $d_{h}=d/h$. The scaled dot-product attention is
	\begin{equation}
		\mathrm{Attn}(\mathbf{Q},\mathbf{K},\mathbf{V})
		= \mathrm{softmax}\!\left(\frac{\mathbf{Q}\mathbf{K}^{\top}}{\sqrt{d_{h}}}\right)\mathbf{V},
	\end{equation}
	extended to multi-head form by per-head linear projections $\mathbf{W}_{Q}^{(j)},$ $\mathbf{W}_{K}^{(j)},\mathbf{W}_{V}^{(j)}\in\mathbb{R}^{d\times d_{h}}$ and an output mix $\mathbf{W}_{O}\in\mathbb{R}^{d\times d}$:
	\begin{equation}
		\mathrm{MHA}(\mathbf{Q},\mathbf{K},\mathbf{V})
		= \Bigl[\,\bigoplus_{j=1}^{h}\mathrm{Attn}\!\bigl(\mathbf{Q}\mathbf{W}_{Q}^{(j)},\mathbf{K}\mathbf{W}_{K}^{(j)},\mathbf{V}\mathbf{W}_{V}^{(j)}\bigr)\Bigr]\mathbf{W}_{O}.
	\end{equation}
	The Multi-head Attention Block (MAB) wraps this with a residual connection, layer normalization (LN)~\citep{ba2016layernorm}, and a row-wise feed-forward network $\mathrm{rFF}_\theta$:
	\begin{align}
		\mathbf{H} &= \mathrm{LN}\!\bigl(\mathbf{Q} + \mathrm{MHA}(\mathbf{Q},\mathbf{K},\mathbf{V})\bigr),
		\\
		\mathrm{MAB}(\mathbf{Q},\mathbf{K},\mathbf{V}) &= \mathrm{LN}\!\bigl(\mathbf{H} + \mathrm{rFF}_\theta(\mathbf{H})\bigr).
	\end{align}

	The Set Transformer recovers linear scaling by
    introducing $m\ll n$ learnable \emph{inducing points}
    $\mathbf{I}\in\mathbb{R}^{m\times d}$ and the Induced Set
    Attention Block (ISAB):
    \begin{align}
        \mathbf{H} &= \mathrm{MAB}(\mathbf{I},\mathcal{X},\mathcal{X})
        \in\mathbb{R}^{m\times d},\\[2pt]
        \mathrm{ISAB}(\mathcal{X}) &= \mathrm{MAB}(\mathcal{X},\mathbf{H},\mathbf{H})
        \in\mathbb{R}^{n\times d},
    \end{align}
    which runs in $O(nm)$ and is permutation \emph{equivariant}
    in $\mathcal{X}$~\citep{lee2019settransformer} (stacking ISAB blocks therefore preserves equivariance; permutation invariance is only obtained once the set is pooled to a fixed-size summary, as described next).

	We also used Pooling by Multihead Attention (PMA) in our ML surrogate. A permutation-invariant fixed-length summary of $\mathcal{X}\in\mathbb{R}^{n\times d}$ is obtained by attending $k$ learnable seed vectors $\mathbf{S}\in\mathbb{R}^{k\times d}$ against $\mathcal{X}$. We omit the row-wise feed-forward pre-processing of $\mathcal{X}$ used in the original formulation~\citep{lee2019settransformer}, giving the simplified form
	\begin{equation}
		\mathrm{PMA}_{k}(\mathcal{X}) = \mathrm{MAB}(\mathbf{S},\mathcal{X},\mathcal{X})\in\mathbb{R}^{k\times d}.
		\label{eq:pma}
	\end{equation}
	A subsequent self-attention block (SAB) and a linear head produce the prediction. Because $\mathbf{S}$ does not depend on the input, the composition $f=\mathrm{LinearHead}\circ\mathrm{SAB}\circ\mathrm{PMA}_{k}\circ\mathrm{ISAB}^{(L)}\circ\cdots$ is permutation invariant by construction.

	\section{Methodology}\label{sec:methodology}

    For developing our  model, STAR-XPC, the process included dataset generation, feature engineering, designing the set transformer architecture, and the cross-validation protocol.

	\subsection{Data Generation Pipeline}

	We used an automated PENELOPE/penEasy pipeline to generate the dataset. Scripts generate a directory tree for all simulations, the input file, the geometry file, and the material file for every composite, then launch and aggregate the simulation runs. The incident photon spectra are four fixed, standardized beam qualities, each generated once
    with the XCOMP5R software~\citep{nowotny1985xcomp5r} and shared by all composites: the ISO $\text{4037-1}$
    narrow-spectrum quality N-30 at \SI{30}{\kilo\volt}~\citep{iso4037}, and the IEC 61267 diagnostic
    qualities RQR~6, RQR~8 and RQR~9 at \SI{80}{\kilo\volt}, \SI{100}{\kilo\volt} and
    \SI{120}{\kilo\volt}, respectively~\citep{iec61267}.

    The polymer used in all samples is polyethylene (C$_{2}$H$_{4})_n$ with a density of
	$\rho_{\mathrm{poly}}=0.95$~g/cm$^{3}$. It is represented as a single token (labeled \texttt{C2H4} in the dataset and SMILES \texttt{C=C} in the
	embedding pipeline). Since the polymer is fixed in all samples, its embedding acts as a constant indicator vector. Hence, the physical contribution of the polymer enters the model only through the bulk descriptors mentioned below and its mass fraction. The shielding materials in this study consist of inorganic compounds
	$\{\mathrm{Bi}_{2}\mathrm{O}_{3},\mathrm{BaSO}_{4},\mathrm{BN},
	\mathrm{B}_{4}\mathrm{C},\mathrm{WO}_{3}\}$ and elemental shielders
	$\{\mathrm{Sb},\mathrm{Sn},\mathrm{Bi},\mathrm{W},\mathrm{Cu},
	\mathrm{B},\mathrm{Ba}\}$, for a total of twelve distinct
	shielding materials. For each sample, the polymer mass fraction is drawn from the
	discrete uniform distribution over a fixed list and the
	remaining mass $1-r_{\mathrm{poly}}$ is split among $K$ (up to three)
	materials via a flat Dirichlet, which ensures that fractions are always positive.

	\begin{gather}
		r_{\mathrm{poly}}\sim\mathcal{U}\{0.20,\,0.30,\,0.40\}, \notag\\
		(r_{1},\ldots,r_{K}) \,\sim\,
		(1-r_{\mathrm{poly}})\cdot\mathrm{Dir}(\mathbf{1}_{K}),\\
		r_{\mathrm{poly}} + \sum_{i=1}^{K} r_{i} = 1. \notag
		\label{eq:sampling}
	\end{gather}
	The bulk mass density of the mixture is computed by the additivity of volumes rule
	\begin{equation}
		\rho \;=\; \biggl(\sum_{i} \frac{r_{i}}{\rho_{i}}\biggr)^{\!-1},
		\label{eq:rho}
	\end{equation}
	and the mean excitation energy is delegated to PENELOPE's \texttt{material} utility, which applies the Bragg additivity rule~\citep{salvat2019penelope}
	\begin{equation}
		\ln I^{exc} \;=\; \frac{\sum_{i} n_{i}\,Z_{i}\,\ln I_{i}^{exc}}{\sum_{i} n_{i}\,Z_{i}},
		\label{eq:bragg}
	\end{equation}
	with $n_{i}$ the atomic number density (atoms/cm$^{3}$) and $Z_{i}$ the atomic number of element $i$, and tabulated elemental $I_{i}^{exc}$ from PENELOPE's database.

    For each configuration, the target variable, the energy transmission ratio, $y=I(t)/I_{0}\in(0,1)$ is computed by normalizing the transmitted energy against the source spectrum integral. We discard configurations that fail the sanity check $y=I(t)/I_{0}\in(0,1)$, are missing bulk physical descriptors, or have no valid additive-identity entry. After this pre-processing, the dataset contains $n=4697$ valid samples. The dataset spans y-values of heavily attenuated and nearly transparent (polymer dominated) samples with a median of 0.539.

	\subsection{Input Feature Engineering and Representation}

	\subsubsection{Molecular Embeddings}

	Each unique molecular constituent is mapped to a SMILES string~\citep{weininger1988smiles} through a deterministic local lookup that handles both organic species (e.g., $\mathrm{C_{2}H_{4}}\!\mapsto\!\mathtt{C=C}$) and inorganic compounds (e.g., $\mathrm{Bi_{2}O_{3}}\!\mapsto\!\mathtt{O=[Bi]O[Bi]=O}$). Pure elements are encoded by their canonical bracketed SMILES ($\mathtt{[W]}$, $\mathtt{[Bi]}$, etc.).

    Each SMILES string $s_{i}$ is tokenized (byte-pair encoding, max length $T=64$)
    and fed through a ChemBERTa model \newline
    (\texttt{seyonec/ChemBERTa-zinc-base-v1})~\citep{chithrananda2020chemberta,sterling2015zinc15},
     with $L=6$ hidden layers of width $768$.
    Instead of collapsing the sequence onto a single \texttt{[CLS]} token,
    we average the four upper hidden layers ($\ell=L-3,\ldots,L$), following the choice of layers used in the feature-based ablation of \citet{devlin2019bert} (which pools them by concatenation or summation, we instead use the mean), then
    take a mask-weighted mean over tokens as follows:
    \begin{equation}
        \bar{\mathbf{H}}_{i} = \frac{1}{4}\sum_{\ell=L-3}^{L} \mathbf{H}^{(\ell)}_{i}
        \in \mathbb{R}^{T\times 768},
        \label{eq:layeravg}
    \end{equation}
    \begin{equation}
        \tilde{\mathbf{e}}_{i} = \frac{\sum_{t=1}^{T} m_{i,t}\,\bar{\mathbf{H}}_{i,t,:}}
        {\max(\sum_{t=1}^{T} m_{i,t},\,\epsilon)} \in \mathbb{R}^{768},
        \label{eq:meanpool}
    \end{equation}
    where $\mathbf{m}_{i}$ is the attention mask and $\epsilon=10^{-9}$ a
    numerical floor. While mask-weighted pooling treats all chemically significant tokens identically, regardless of where they are in the SMILES string, layer-averaging collects a combination of syntactic and semantic information that is not available from any single layer. Finally, we use $\ell_{2}$ normalization
    to remove norm-induced scale heterogeneity between organic and
    inorganic species.

	\subsubsection{Composition Features}

	A single composite is represented as a fixed-size set of $N=11$ component slots (up to 10 material slots plus the polymer slot). Each non-empty slot $i$ carries the constituent's molecular embedding $\mathbf{e}_{i}$ and its mass fraction $r_{i}$, renormalized so that $\sum_{i} r_{i}=1$. Empty slots receive a zero embedding and $r_{i}=0$, but, like every other slot, they also carry the global descriptor vector $\mathbf{g}$ (Section~\ref{sec:global}). Empty slots are not masked in attention: all padding tokens are identical, so they act as a repeated global-descriptor token that the encoder and the PMA block can attend to. Because the padding tokens are identical, the model remains exactly invariant to any permutation of the slots. However, the number of padding tokens is part of the input, so the trained network is specific to $N=11$.

	\subsubsection{Global Physical Parameters}\label{sec:global}
    Four bulk descriptors are recorded per composite: the tube voltage $V_{in}$, which indexes one of the four standard beam qualities
    (one of $\{30,80,100,120\}$~kV), the bulk mass density $\rho$ (g/cm$^{3}$), the slab
    thickness $t$ (mm), and the Bragg-additivity mean excitation energy $I^{exc}$ (eV).
    We note that $I^{exc}$ is a charged-particle stopping-power parameter and is not directly related to the photon attenuation in the 30-120~kV range. Moreover, the secondary-electron transport has
    a negligible effect on the transmitted energy for the slab thicknesses considered here.
    We retain it purely as an engineered compositional summary. By Eq.~\eqref{eq:bragg},
    $\ln I^{exc}$ is an $n_iZ_i$-weighted logarithmic mean over the constituent elements
    and therefore acts as a smooth scalar proxy for the high-$Z$ content of the mixture.
    Since it is a deterministic function of the constituents and their mass fractions, it
    carries no information beyond the token set. In an ablation study, removing $I^{exc}$ as an input increased MAE in 9 of 10 folds.

    Each descriptor is standardized by Z-normalization,
	\begin{equation}
		x_{\mathrm{norm}} = \frac{x-\mu_{x}}{\sigma_{x}},
	\end{equation}
	where $\mu_{x},\sigma_{x}$ are computed exclusively on the training partition of each cross-validation fold (Section~\ref{sec:cv}).
	The single-run model used in Sections~\ref{sec:baselines}, \ref{sec:cvperf} and~\ref{sec:generalization} is trained with $\mu_x,\sigma_x$ computed on its 80\% training split and is evaluated with the same statistics. The standardized descriptor input vector is
	\begin{equation}
		\mathbf{g} \;=\; [V_{\mathrm{norm}},\,\rho_{\mathrm{norm}},\,t_{\mathrm{norm}},\,I_{\mathrm{norm}}]\in\mathbb{R}^{4}.
	\end{equation}
	Including $V$ as an explicit descriptor is what allows a single network to handle the four spectrally distinct beam settings without a per-voltage head. The vector $\mathbf{g}$ is concatenated to all component tokens, so that every element of the input set carries complete information about the global geometry, beam quality, and bulk material state. The final per-component input is the concatenation of the $768$-dimensional molecular embedding, the scalar mass fraction, and the four global descriptors,
	\begin{equation}
		\mathbf{x}_{i} \;=\; \bigl[\mathbf{e}_{i};\,r_{i};\,\mathbf{g}\bigr] \in \mathbb{R}^{773}.
	\end{equation}

    \begin{figure*}[pos=!t]
		\centering
		\includegraphics[width=\textwidth]{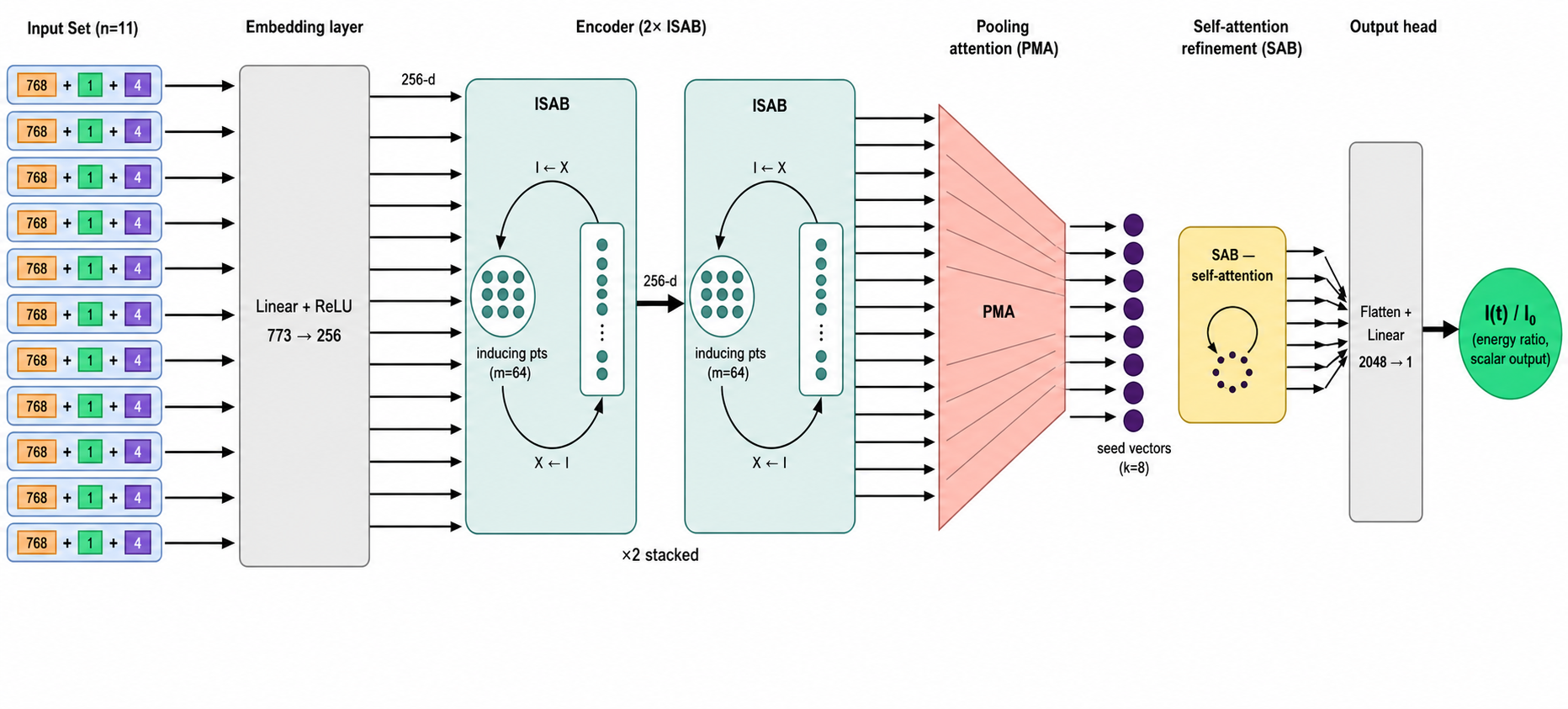}
		\caption{Set Transformer architecture of STAR-XPC. Each of the $N=11$ constituent tokens is the concatenation of its $768$-dimensional ChemBERTa embedding, its scalar mass fraction, and the four global descriptors $\mathbf{g}$. The token set is processed by two permutation-equivariant ISAB encoder blocks, pooled by a PMA block into $k=8$ seed vectors, mixed by a self-attention block (SAB), and mapped by a linear head to the scalar energy transmission ratio $\hat{y}$.}
		\label{fig:architecture}
	\end{figure*}
	\subsection{Set Transformer Architecture}\label{sec:arch}

    \subsubsection{Core Architecture}

    The network outputs a scalar prediction $\hat{y}\in\mathbb{R}$ which is regressed onto our target, the energy transmission ratio $y\in(0,1)$, after taking the padded set $\mathbf{X}=\{\mathbf{x}_{1},\ldots,\mathbf{x}_{11}\}\in\mathbb{R}^{11\times 773}$ as an input. The architecture (Figure~\ref{fig:architecture}) is a simple implementation of the Set Transformer of \citet{lee2019settransformer}:

	\begin{align}
		\mathbf{Z}_{0} &= \mathrm{ReLU}\bigl(\mathbf{X}\,\mathbf{W}_{e} + \mathbf{b}_{e}\bigr),
		\\
		\mathbf{Z}_{\ell} &= \mathrm{ISAB}_{m}^{(\ell)}\bigl(\mathbf{Z}_{\ell-1}\bigr),\quad \ell=1,2,
		\\
		\mathbf{P} &= \mathrm{PMA}_{k}\bigl(\mathbf{Z}_{2}\bigr),
		\quad
		\mathbf{Q} = \mathrm{SAB}\bigl(\mathbf{P}\bigr),
		\\
		\hat{y} &= \mathbf{w}_{o}^{\top}\mathrm{vec}(\mathbf{Q}) + b_{o}.
	\end{align}

    Every MAB, and hence every ISAB, PMA and SAB block, contains a per-element (token-wise) feed-forward sub-network, which we instantiate as three \texttt{Linear-ReLU} layers acting independently on each token (the $\mathrm{rFF}$ block in the notation of \citet{lee2019settransformer}).    The hyperparameters used in this work, summarized in Table~\ref{tab:hparams}, were selected by testing alternative configurations.

	\begin{table}[pos=!htb]
		\centering
		\caption{Set Transformer hyperparameters are decided through extensive testing. The resulting network has $3{,}004{,}417$ trainable parameters.}
		\label{tab:hparams}
		\begin{tabular}{lc}
			\toprule
			Hyperparameter & Value \\
			\midrule
			Hidden dimension $d$ & 256 \\
			Attention heads $h$ & 32 \\
			ISAB inducing points $m$ & 64 \\
			PMA seed vectors $k$ & 8 \\
			ISAB layers & 2 \\
			Row-wise FF layers per block & 3 \\
			Input dimension & 773 \\
			Output dimension & 1 \\
			Maximum set cardinality & 11 \\
			\bottomrule
		\end{tabular}
	\end{table}

    On a CPU (no GPU acceleration), a forward pass measures $2.6$~ms per sample at batch size $1$ and $0.99$~ms per sample at batch size $256$, both remain several orders of magnitude below the multi-hour PENELOPE/penEasy reference runs regardless of hardware.

	\subsubsection{Training Configuration}

	The model is trained by minimizing the mean squared error
	\begin{equation}
		\mathcal{L}(\theta) \;=\; \frac{1}{B}\sum_{j=1}^{B}\bigl(y_{j}-\hat{y}_{j}(\theta)\bigr)^{2}
		\label{eq:mse}
	\end{equation}
	with $B=256$, using the AdamW optimizer~\citep{loshchilov2019adamw} at base learning rate $\alpha_{0}=10^{-3}$ and weight decay $\lambda=10^{-5}$. We deliberately regress the raw target $y$ under a squared-error loss rather than a logit- or log-transformed target. The learning rate follows a cosine annealing schedule, without the warm restarts of \citet{loshchilov2017sgdr}, with period $T_{\max}=200$~epochs,
	\begin{equation}
		\alpha_{e} \;=\; \frac{\alpha_{0}}{2}\!\left[1+\cos\!\left(\frac{\pi e}{T_{\max}}\right)\right],
		\qquad T_{\max} = 200,
	\end{equation}
	so that the schedule completes a full half-cosine over $200$ epochs. Gradient norms are clipped to $1.0$ to stabilize the early epochs. Training runs for at most $200$ epochs with early stopping after $50$ consecutive epochs without validation-loss improvement, so the schedule could in principle be cut short; in practice, every one of the ten cross-validation folds (Section~\ref{sec:cv}) and the single-run training reported in this paper ran the full $200$ epochs without early stopping firing.

	\subsection{Cross-Validation Protocol}\label{sec:cv}

    We estimate generalization with 10-fold cross validation~\citep{kohavi1995cv}, jointly stratified on chemistry and target value. To avoid leakage~\citep{kaufman2012leakage}, all normalization statistics are fit on the training rows of each fold only (see below). The combined stratification method is indeed motivated by two factors: the chemistry of the dataset not being uniform and the energy transmission ratio target spanning several decades. Some chemistries may be overrepresented by folds if we just define it on $y$, while folds defined solely on chemistry may collapse $y$ into a tight region.

    Hence, we decided to stratify based on the Cartesian product of (i) main shielding material and (ii) empirical-quantile bin of the target. Let $\mathrm{mol}_{i}^{\star}\!=\!\arg\max_{j}r_{ij}$ denote the highest-mass-fraction non-polymer constituent in sample $i$. Given a desired number of energy bins $B_{E}$, define $\mathrm{bin}_{E}(y)\in\{0,...,B_{E}\!-\!1\}$ via the empirical-quantile boundaries $\hat{F}^{-1}(b/B_{E})$ for $b=1,...,B_{E}\!-\!1$. The joint stratum label is
	$s_{i} = \bigl(\mathrm{bin}_{E}(y_{i}),\;\mathrm{mol}_{i}^{\star}\bigr)$.

	$B_{E}$ is adaptively chosen as the largest value keeping at least 10 samples per stratum (\texttt{StratifiedKFold}~\citep{pedregosa2011scikit}), ensuring every fold gets at least one sample from each stratum, with each fold using a 90/10 train+validation/test split (and ~89/11 train /validation within that when early stopping is used), while the separate single-run training explained in Section~\ref{sec:methodology} uses its own 80/20 train/validation random split. Please note that different ratio compositions correspond to different material compositions and no material composition is seen twice in the dataset. Therefore, each unique composition is tested with a single thickness, except for the generalization analysis at the end of the performance evaluation with held-out compositions.

	The four global descriptors are applied unaltered to the validation and test partitions after being Z-scored using statistics that are only calculated on the training rows of each fold. This is crucial since it would boost the perceived accuracy without altering the underlying generalization if validation/test statistics were leaked. We present root mean squared error (RMSE), mean squared error (MSE), mean absolute error (MAE), and the normalized-MSE percentage (NMSE\%).
	\begin{equation}
		\mathrm{NMSE\%} \;=\; 100\times\frac{\mathrm{MSE}}{\mathbb{E}[y_{\mathrm{true}}^{2}]}.
		\label{eq:nmse}
	\end{equation}

	\section{Results and Discussion}\label{sec:results}

    \subsection{Baseline Comparison}\label{sec:baselines}

	We argued that fixed-length, order-dependent regressors are a poor fit for this problem. We compare the Set Transformer against three standard regressors given the exact same information, under the same 10-fold protocol as Section~\ref{sec:cv}. Each baseline gets the same molecular embeddings, mass fractions, and global descriptors as the Set Transformer (STAR-XPC), just flattened into one fixed-length vector (the 11 slot embeddings, then the 11 mass fractions, then the 4 global descriptors, $8463$ numbers total) instead of being read as a set.

    We tested Ridge regression~\citep{hoerl1970ridge} ($\ell_2$-penalty) $\alpha$ tuned on held-out data (please note that the plain linear regression cannot even be fit here, since there are more features than training rows), Random Forest~\citep{breiman2001randomforests}, and RBF-kernel support vector regression (SVR)~\citep{drucker1996svr}. All hyperparameters are selected with \texttt{scikit-learn}~\citep{pedregosa2011scikit} on the validation split of each fold.

    %%%%%%%%%%%%%%%%%%%%%%%%%%%%%%%%%%%%%%%%%%%%%%%%%%%%%%
	  \begin{figure}[pos=!htb]
		\centering
		\begin{subfigure}{0.49\linewidth}
			\includegraphics[width=\linewidth]{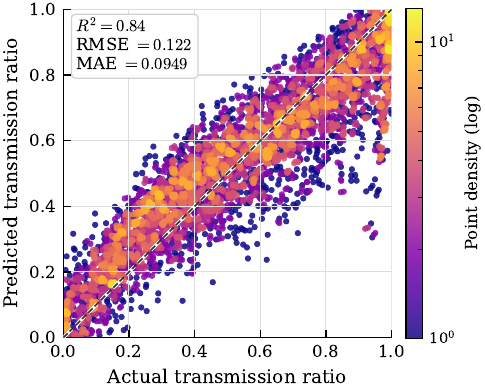}
			\caption{Ridge Regression}
		\end{subfigure}\hfill
		\begin{subfigure}{0.49\linewidth}
			\includegraphics[width=\linewidth]{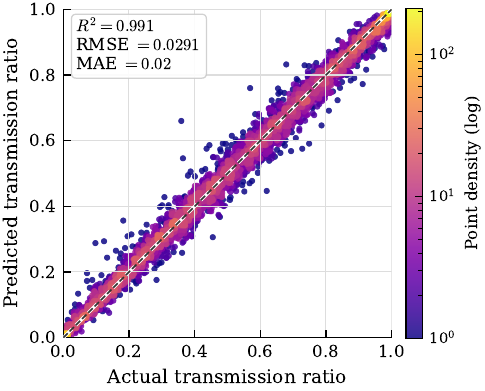}
			\caption{Random Forest}
		\end{subfigure}\\[0.5em]
		\begin{subfigure}{0.49\linewidth}
			\includegraphics[width=\linewidth]{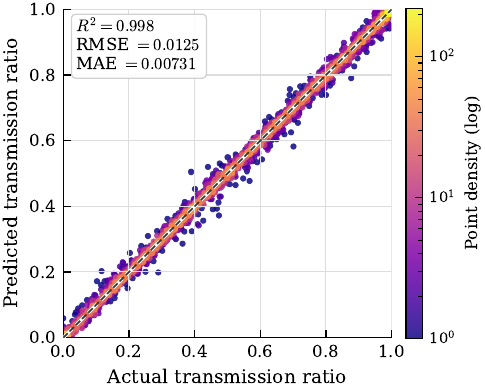}
			\caption{SVR (RBF)}
		\end{subfigure}\hfill
		\begin{subfigure}{0.49\linewidth}
			\includegraphics[width=\linewidth]{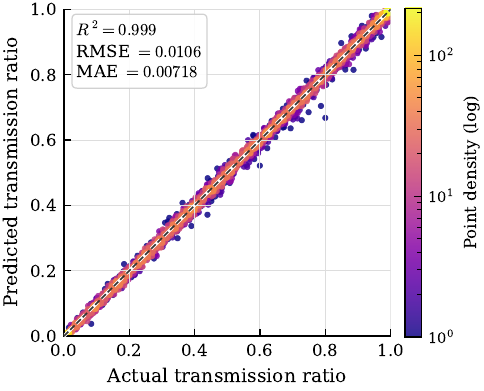}
			\caption{STAR-XPC}
		\end{subfigure}
		\caption{Predicted vs.\ simulated energy transmission ratios for each of the three baselines and STAR-XPC, aggregated over the out-of-fold predictions of the identical 10-fold cross-validation ($n=4697$).}
		\label{fig:baseline_parity}
	\end{figure}
    %%%%%%%%%%%%%%%%%%%%%%%%%%%%%%%%%%%%%%%%%%%%%%%%%%%%
    Figure~\ref{fig:baseline_parity} shows the aggregate parity plots (ground truth vs.\ estimate) for the three baselines and STAR-XPC, and Table~\ref{tab:baselines_summary} lists the corresponding NMSE\% values. Two things stand out. First, Ridge regression is far worse than every other model ($R^2=0.84$ in-distribution, versus $\ge 0.99$ for the others), confirming that the attenuation problem is nonlinear. Second, in the standard 10-fold comparison SVR comes close to STAR-XPC ($\mathrm{NMSE\%}=0.0312\%$ vs.\ $0.0278\%$ on $80$-\SI{120}{\kilo\volt}). We attribute the competitiveness of SVR to the in-distribution setting: the embedding cache contains only $13$ distinct molecule vectors, so an RBF kernel can effectively memorize ``similar known composition $\Rightarrow$ similar output'', and every fold's test samples use the same fixed column order as training, so the flattened input is never penalized for depending on it. The two experiments below test these two advantages.
    %%%%%%%%%%%%%%%%%%%%%%%%%%%%%%%%%%%%%%%%%%%%%%%%%%
    \begin{table}[pos=!htb]
		\centering
		\caption{NMSE\% in-distribution (10-fold CV, out-of-fold predictions restricted to $80$-\SI{120}{\kilo\volt}, $n=3996$) versus the mean over the six held-out unseen-combination test sets, and the ratio between them.}
		\label{tab:baselines_summary}
		\footnotesize
		\begin{tabular}{lccc}
			\toprule
			Model & CV NMSE\% & Held-out NMSE\% & Ratio \\
			\midrule
			Ridge (linear)  & 3.25   & 7.89   & $2.4\times$ \\
			Random Forest   & 0.174  & 0.433  & $2.5\times$ \\
			SVR (RBF)       & 0.0312 & 0.0804 & $2.6\times$ \\
			STAR-XPC        & 0.0278 & 0.0496 & $\mathbf{1.8\times}$ \\
			\bottomrule
		\end{tabular}
	\end{table}
    %%%%%%%%%%%%%%%%%%%%%%%%%%%%%%%%%%%%%%%%%%%%%%%%%%%%%%%

    In Table~\ref{tab:baselines_summary}, the two NMSE\% columns come from different training regimes. The CV column, like Figure~\ref{fig:baseline_parity}, aggregates the out-of-fold predictions of the ten cross-validation models of each method, whereas the held-out column is obtained from a single training run per method on the full training pool, evaluated once on the six held-out compositions. Since the held-out compositions are restricted to the $80$-\SI{120}{\kilo\volt} samples ($n=3996$), which is why the STAR-XPC entry is lower than the all-voltage $\mathrm{NMSE\%}=0.0294\%$ reported in Section~\ref{sec:cvperf}.

    The held-out column of Table~\ref{tab:baselines_summary} is a more informative comparison related to generalization. There, STAR-XPC reaches a mean NMSE\% of $0.0496\%$ against $0.0804\%$ for SVR. From CV to held-out, the three flattened baselines degrade by a similar $2.4$-$2.6\times$ despite in-distribution errors spanning two orders of magnitude, whereas STAR-XPC degrades by $1.8\times$. What separates the two models is invariance to input order.

	Another key point is that every fold's test set preserves the fixed column order as training, risking that the flattened baselines memorize this input order rather than learning dependence. A flattened vector has no notion that "slot 1" and "slot 2" are interchangeable and that ordering does not matter, while the Set Transformer's set input does, by construction. To illustrate this even further, we took one sample per held-out composition and generated 40 random reorderings of its 11 molecule slots and re-ran all three trained models on every version.
    %%%%%%%%%%%%%%%%%%%%%%%%%%%%%%%%%%%%%%%%%%%%%%%%%%%%%%%%
	\begin{figure*}[pos=!t]
		\centering
		\includegraphics[width=\linewidth]{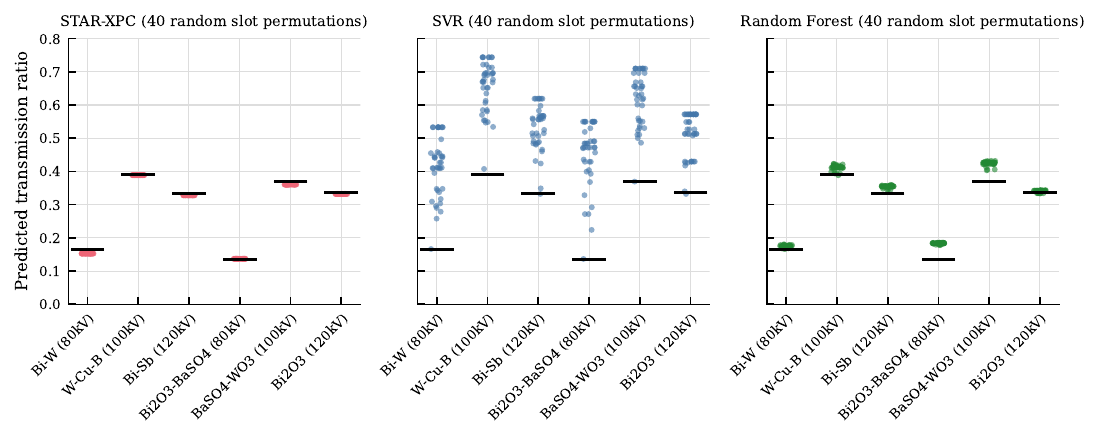}
		\caption{Predicted transmission ratio across 40 random slot reorderings of the same physical sample (black tick: ground truth), all three panels on a common $y$-scale. STAR-XPC's predictions collapse to a single point per case, SVR's scatter across most of the target range, and the Random Forest's, although far less sensitive than SVR's, still fail to collapse to a single value per sample.}
		\label{fig:permutation_stress}
	\end{figure*}

	In Figure~\ref{fig:permutation_stress}, STAR-XPC's predictions do not move (standard deviation $\le 1.3\times 10^{-7}$, i.e.\ floating-point noise), while the standard deviation of predicted transmission ratios for SVR is between 0.067 and 0.098 for the identical physical sample, simply because the molecules were listed in a different order. The Random Forest lies between these two extremes. Its standard deviation stays within $0.0018$-$0.0078$, since its axis-aligned splits rest largely on the four global descriptors, which are per-sample and therefore untouched by the reordering. It is nevertheless not order-invariant; over the 40 reorderings of a single sample its prediction still moves by up to $0.034$, roughly $9\%$ of that sample's transmission ratio. In other words, the apparent superiority of the flattened baselines on in-distribution test data may not reflect genuine physical understanding, it instead stems from the models memorizing the fixed order in which input columns happen to be arranged in the data. STAR-XPC, on the other hand, is unaffected by these random permutations, since it has no fixed input order to memorize in the first place.

	In short, a well-tuned kernel method performs close to STAR-XPC when interpolating compositions it has effectively already seen, but is outperformed on new combinations (mean held-out NMSE\% $0.0804\%$ vs.\ $0.0496\%$) and is not robust to input reordering. STAR-XPC is the only model here that is accurate, generalizes to new combinations, and is provably indifferent to how the input happens to be listed.

	\subsection{Detailed Cross-Validation Performance of STAR-XPC}\label{sec:cvperf}

\begin{figure}[pos=!htb]
		\centering
		\includegraphics[width=\linewidth]{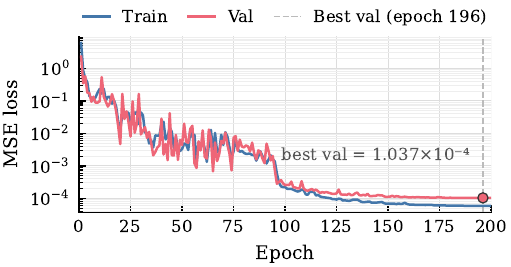}
		\caption{Training and validation MSE loss for the single-run model. The grey dashed line marks the best-validation epoch and the red dot marks its loss value.}
		\label{fig:train_val_loss}
	\end{figure}

    Figure~\ref{fig:train_val_loss} shows the training and validation loss plots. Training and validation loss both decrease rapidly in the first $\sim\!30$ epochs and reach a plateau at the $\mathcal{O}(10^{-4})$ level. Since the two curves closely follow each other throughout the training phase, we may deduce that we didn't encounter overfitting.
	%%%%%%%%%%%%%%%%%%%%%%%%%%%%%%%%%%%%%%%%%
    %%%%%%%%%%%%%%%%%%%%%%%%%%%%%%%%%%%%%%%%%

    When tested on the whole pool of data $(n=4697)$, STAR-XPC exhibits high prediction accuracy over all 10 cross-validation folds. Aggregated across folds, we observe $\overline{\mathrm{MSE}} = (1.12\pm 0.56)\times 10^{-4}$ and $\overline{\mathrm{MAE}} = 0.00718\pm 0.00148$ with $R^{2} = 0.9988$ (mean $\pm$ sample standard deviation over folds).

    The aggregate metrics above are residual-scale statistics that we used for the training. For the performance evaluation, we complement them with a per-sample relative measure, we additionally report the per-sample absolute percentage error (APE),
	\begin{equation}
		\mathrm{APE}_{i} \;=\; 100\cdot\frac{\lvert\hat{y}_{i}-y_{i}\rvert}{\max \bigl(\lvert y_{i}\rvert,\,\epsilon\bigr)},\qquad \epsilon = 10^{-9},
		\label{eq:ape}
	\end{equation}
	mean APE (MAPE), and an $\alpha$-trimmed mean APE ($\mathrm{MAPE}_{\alpha}$)
	\begin{equation}
		\mathrm{MAPE}_{\alpha} \;=\; \frac{1}{N(1-2\alpha)}\sum_{i\in\mathcal{I}_{\alpha}} \mathrm{APE}_{i},\qquad \alpha = 0.10,
		\label{eq:trimmedmape}
	\end{equation}
	where $\mathcal{I}_{\alpha}$ is the index set of the central $80\%$ of the $\mathrm{APE}_{i}$ values (i.e.\ the lowest and highest $10\%$ are discarded). In this case, trimming is quite needed due to the fact that samples with a transmission ratio $y$ near numerical zero (i.e., $\hat{y}$ even with a very small error) produce a non-representative, excessively high percentage error that dominates the MAPE without providing any insight about  accuracy.

    Pooled over the out-of-fold predictions of all ten folds, we observe $\mathrm{MAPE} = 203\%$ due to the near-zero-$y$ samples. This figure is driven almost entirely by the $129$ near-opaque samples with $y<0.01$ ($2.7\%$ of the pool). Please note that their absolute errors are small, but the APE denominator vanishes as $y\to 0$. The quantiles of the per-sample APE distribution, $(P_{50},P_{75},P_{90},P_{95})=(1.15\%,\,2.84\%,\,6.75\%,\,12.04\%)$, make this concentration explicit: the median and the trimmed mean are of the same order, while the untrimmed mean is dominated by the extreme upper tail. Discarding the lowest and highest $10\%$ of the APE values results in $\mathrm{MAPE}_{0.10} = 1.66\%$ over all four tube voltages and $\mathrm{MAPE}_{0.10} = 1.32\%$ over $80$-\SI{120}{\kilo\volt} ($n=3996$). Therefore, $\mathrm{MAPE}_{0.10}$ is included as a more representative and practitioner-friendly per-sample relative-error metric for non-opaque configurations.

	\subsection{Analysis of STAR-XPC}\label{sec:robustness}

    For evaluating model robustness, we analyzed the effect of energy-level and thickness. Even if the \SI{30}{\kilo\volt} tube voltage is not so common in practical applications, the \SI{30}{\kilo\volt} is included in training in order to assist the network to learn generalization across the entire kV range, and in the all-voltage CV metrics. Given that the model is cross-validated, the overall cross-validation figures of Section~\ref{sec:results} above are reported over all four voltages. Nonetheless, starting with this subsection, we report every stratified metric on the $80$, $100$, and \SI{120}{\kilo\volt} data ($n=3996$).

	\subsubsection{Energy-level performance}

    As expected, the beam spectrum has a direct relation with the performance (Table~\ref{tab:energy_performance}). The representative relative error decreases monotonically with tube voltage, from $\mathrm{MAPE}_{0.10}=1.65\%$ at \SI{80}{\kilo\volt} to $1.25\%$ at \SI{100}{\kilo\volt} and $1.13\%$ at \SI{120}{\kilo\volt}, since harder spectra are attenuated less and the mean target rises accordingly ($\overline{y}=0.495$, $0.556$, $0.703$). The absolute and squared-error metrics remain small in all three regimes (MAE $\le 0.0087$, $\mathrm{NMSE\%}\le 0.032\%$) but do not follow a monotonic trend, because MAE scales with the magnitude of $y$ while NMSE\% is normalised by $\mathbb{E}[y^{2}]$ on each subset.
    %%%%%%%%%%%%%%%%%%%%%%%%%%%%%%%%%%%%%%%%%%%%
	\begin{table}[pos=!htb]
		\centering
		\caption{Performance stratified by tube voltage ($80$-\SI{120}{\kilo\volt}, $n=3996$)}
		\label{tab:energy_performance}
		\footnotesize
		\begin{tabular}{lccccc}
			\toprule
			Tube vltg. & $n$ & MAE & NMSE\% & $\mathrm{MAPE}_{0.10}$  & $\overline{y}$ \\
			\midrule
			80 kV  & 1367 & 0.00683 & 0.032 &  1.65 & 0.495 \\
			100 kV & 1318 & 0.00673 & 0.023 &  1.25 & 0.556 \\
			120 kV & 1311 & 0.00865 & 0.028 &  1.13 & 0.703 \\
			\bottomrule
		\end{tabular}
	\end{table}

	\subsubsection{Thickness-dependent accuracy}\label{sec:thickness}

	Our data-generation pipeline sweeps single-element shielding materials (B, Ba, Bi, Cu, Sb, Sn, W) only up to $1.5$~mm, while molecular or compound fillers (B$_{4}$C, BN, BaSO$_{4}$, Bi$_{2}$O$_{3}$, WO$_{3}$) are swept up to $2.5$~mm, and the mix of the two classes is not constant across thickness bins.

    In Table~\ref{tab:thickness_mae_elements} and \ref{tab:thickness_mae_compounds}, performance is stratified by thickness on single-element and compound cases. $\mathrm{MAPE}_{0.10}$ and NMSE\% increase monotonically as thickness increases, while MAE has no trend. This is the expected first-order behavior of the transmission ratio $y=I(t)/I_{0}$. Writing $y(t)=e^{-\mu t}$ for an effective attenuation coefficient $\mu$, an error $\Delta\mu$ propagates as
\begin{equation}
    \frac{|\Delta y|}{y} \;=\; t\,|\Delta\mu|,
    \qquad
    |\Delta y| \;=\; t\,|\Delta\mu|\,e^{-\mu t},
    \label{eq:bl_sensitivity}
\end{equation}
so the relative error scales linearly with $t$ whereas the absolute error stays bounded by the decaying exponential, explaining the flat MAE and the increasing $\mathrm{MAPE}_{0.10}$ reported here.

	\subsubsection{Chemistry-stratified performance}

    Table~\ref{tab:dominant_mae} stratifies MAE, NMSE\%, and $\mathrm{MAPE}_{0.10}$ by the main shielding material, and the table is sorted by $\mathrm{MAPE}_{0.10}$. STAR-XPC estimates the compound constituents more accurately than the elemental ones. The six worst rows by $\mathrm{MAPE}_{0.10}$ (Sb, Sn, Cu, Bi, Ba, W) are all single elements, and the lowest MAE ($0.0046$-$0.0048$) is attained by BN, B$_{4}$C and BaSO$_{4}$. The clearest evidence comes from the four elements that appear both in elemental and in compound form: in every such pair the compound has both the lower MAE and the lower $\mathrm{MAPE}_{0.10}$ (Bi $2.43\%$ vs. Bi$_{2}$O$_{3}$ $1.19\%$; W $2.58\%$ vs. WO$_{3}$ $1.23\%$; Ba $2.44\%$ vs. BaSO$_{4}$ $0.73\%$; B $1.24\%$ vs. BN $0.42\%$ and B$_{4}$C $0.47\%$). Sample count accounts for this only in part: across the twelve materials MAE decreases with $n$ (Spearman $\rho=-0.70$, Pearson $r=-0.64$), yet WO$_{3}$ ($n=251$) is estimated more accurately than W ($n=289$) despite having fewer samples. One possible reason is that, at the same mass fraction, an elemental constituent puts more high-$Z$ mass into the slab than its compound counterpart. This leads to stronger attenuation with more spectral structure. We leave a controlled test of this hypothesis to future work.
    %%%%%%%%%%%%%%%%%%%%%%%%%%%%%%%%%%%%%%%%%
    \begin{table}[pos=!htb]
		\centering
		\caption{Performance stratified by thickness bins for samples whose main material is a single element (B, Ba, Bi, Cu, Sb, Sn, W, $80$-\SI{120}{\kilo\volt}, $n=1941$).}
		\label{tab:thickness_mae_elements}
		\footnotesize
		\begin{tabular}{lcccc}
			\toprule
			Thickness bin (mm) & $n$ & MAE & NMSE\% & $\mathrm{MAPE}_{0.10}$  \\
			\midrule
			$(0.149,\,0.45]$ & 499 & 0.01031 & 0.0294 & 1.25 \\
			$(0.45,\,0.75]$  & 454 & 0.00932 & 0.0440 & 1.58 \\
			$(0.75,\,1.05]$  & 442 & 0.01005 & 0.0707 & 2.29 \\
			$(1.05,\,1.5]$   & 546 & 0.01056 & 0.1196 & 3.32 \\
			\bottomrule
		\end{tabular}
	\end{table}
    %%%%%%%%%%%%%%%%%%%%%%%%%%%%%%%%%%%%%%%%%%
	\begin{table}[pos=!htb]
		\centering
		\caption{Performance stratified by thickness bins for samples whose main material is a compound (B$_{4}$C, BN, BaSO$_{4}$, Bi$_{2}$O$_{3}$, WO$_{3}$, $80$-\SI{120}{\kilo\volt}, $n=2055$).}
		\label{tab:thickness_mae_compounds}
		\footnotesize
		\begin{tabular}{lcccc}
			\toprule
			Thickness bin (mm) & $n$ & MAE & NMSE\% & $\mathrm{MAPE}_{0.10}$  \\
			\midrule
			$(0.199,\,0.4]$ & 558 & 0.00466 & 0.0057 & 0.51 \\
			$(0.4,\,0.75]$  & 433 & 0.00454 & 0.0072 & 0.59 \\
			$(0.75,\,1.5]$  & 631 & 0.00515 & 0.0201 & 0.92 \\
			$(1.5,\,2.5]$   & 433 & 0.00495 & 0.0213 & 1.48 \\
			\bottomrule
		\end{tabular}
	\end{table}
    %%%%%%%%%%%%%%%%%%%%%%%%%%%%%%%%%%%%%%%%%%%%%%%%
    \begin{table}[pos=!htb]
		\centering
		\caption{Per-shielding-material MAE, NMSE\%, and $\alpha$-trimmed MAPE ($80$-\SI{120}{\kilo\volt}, $n=3996$).}
		\label{tab:dominant_mae}
        \footnotesize
		\begin{tabular}{lcccc}
			\toprule
			\shortstack[l]{Main Shielding\\ Material} & $n$ & MAE & NMSE\% & $\mathrm{MAPE}_{0.10}$ \\
			\midrule
			BN              & 371 & 0.0047 & 0.012 & 0.42 \\
			B$_{4}$C        & 309 & 0.0046 & 0.009 & 0.47 \\
			BaSO$_{4}$      & 525 & 0.0048 & 0.009 & 0.73 \\
			Bi$_{2}$O$_{3}$ & 599 & 0.0049 & 0.015 & 1.19 \\
			WO$_{3}$        & 251 & 0.0052 & 0.017 & 1.23 \\
			B               & 248 & 0.0105 & 0.027 & 1.24 \\
			Sb              & 312 & 0.0088 & 0.040 & 1.61 \\
			Sn              & 278 & 0.0092 & 0.041 & 1.80 \\
			Cu              & 197 & 0.0142 & 0.107 & 2.28 \\
			Bi              & 383 & 0.0093 & 0.062 & 2.43 \\
			Ba              & 234 & 0.0113 & 0.065 & 2.44 \\
			W               & 289 & 0.0093 & 0.065 & 2.58 \\
			\bottomrule
		\end{tabular}
	\end{table}
    %%%%%%%%%%%%%%%%%%%%%%%%%%%%%%%%%%%%%%%%%%%%%%

    %%%%%%%%%%%%%%%%%%%%%%%%%%%%%%%%%%%%%%%%%%%%%%%%%%%%%%
	  \begin{figure*}[pos=!t]
		\centering
		\begin{subfigure}[t]{0.32\linewidth}
			\includegraphics[width=\linewidth]{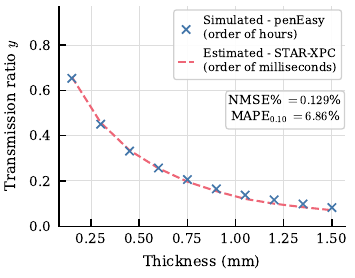}
			\caption{Bi-W at \SI{80}{\kilo\volt};\\$31/39/30$~wt\% Bi/W/(C$_{2}$H$_{4})_n$}
		\end{subfigure}\hfill
		\begin{subfigure}[t]{0.32\linewidth}
			\includegraphics[width=\linewidth]{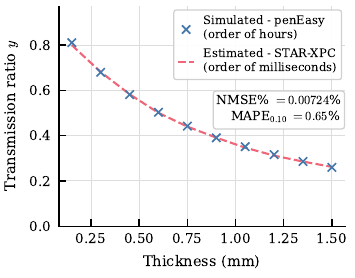}
			\caption{W-Cu-B at \SI{100}{\kilo\volt};\\$48/9/3/40$~wt\% W/Cu/B/(C$_{2}$H$_{4})_n$}
		\end{subfigure}\hfill
        \begin{subfigure}[t]{0.32\linewidth}
			\includegraphics[width=\linewidth]{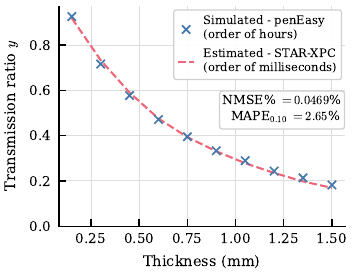}
			\caption{Bi-Sb at \SI{120}{\kilo\volt};\\$60/20/20$~wt\% Bi/Sb/(C$_{2}$H$_{4})_n$}
		\end{subfigure}\\[0.5em]
		\begin{subfigure}[t]{0.32\linewidth}
			\includegraphics[width=\linewidth]{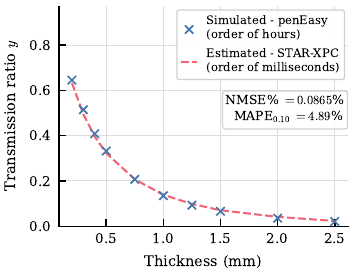}
			\caption{Bi$_{2}$O$_{3}$-BaSO$_{4}$ at \SI{80}{\kilo\volt};\\$40/40/20$~wt\% Bi$_{2}$O$_{3}$/BaSO$_{4}$/(C$_{2}$H$_{4})_n$}
		\end{subfigure}\hfill
        \begin{subfigure}[t]{0.32\linewidth}
			\includegraphics[width=\linewidth]{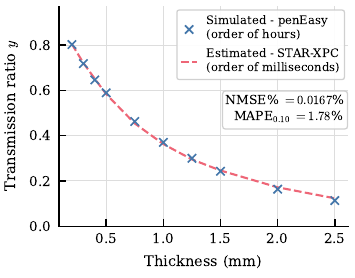}
			\caption{BaSO$_{4}$-WO$_{3}$ at \SI{100}{\kilo\volt};\\$35/35/30$~wt\% BaSO$_{4}$/WO$_{3}$/(C$_{2}$H$_{4})_n$}
		\end{subfigure}\hfill
        \begin{subfigure}[t]{0.32\linewidth}
			\includegraphics[width=\linewidth]{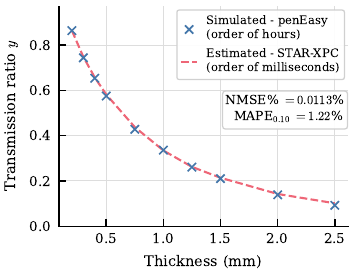}
			\caption{Bi$_{2}$O$_{3}$ at \SI{120}{\kilo\volt};\\$80/20$~wt\% Bi$_{2}$O$_{3}$/(C$_{2}$H$_{4})_n$}
		\end{subfigure}
		\caption{Transmission ratio versus material thickness for held-out data to analyze the generalization capability of STAR-XPC. Each sub-caption gives the tube voltage and the mass fractions of the shielding material and of the polyethylene (C$_{2}$H$_{4})_n$ matrix, which sum to $100$~wt\%.}
		\label{fig:test_curves}
	\end{figure*}
    %%%%%%%%%%%%%%%%%%%%%%%%%%%%%%%%%%%%%%%%%%%%%%%%%%%%

    \subsection{Generalization Capability of STAR-XPC}\label{sec:generalization}\label{sec:pbgd}
    %%%%%%%%%%%%%%%%%%%%%%%%%%%%%%%%%%%%%%%%%%%%%%%%%%%%%%
	To assess generalization beyond the cross-validation pool, we constructed six dedicated test sets corresponding to compositions not seen during training STAR-XPC, each containing $10$ thickness samples spanning the full range. In Figure~\ref{fig:test_curves}, thickness versus transmission ratios of simulated and estimated values are plotted. Moreover, average NMSE\% and $\mathrm{MAPE}_{0.10}$ (\%) for each case are given in subfigures. All six cases achieve $\mathrm{NMSE\%}\le 0.129\%$ while $\mathrm{MAPE}_{0.10} \le 6.86\%$. The two largest $\mathrm{MAPE}_{0.10}$ occur at \SI{80}{\kilo\volt}, Bi-W and Bi$_{2}$O$_{3}$-BaSO$_{4}$, since transmission values are small and even a small error is large in terms of percentage error. Overall, the proposed model generalizes compositions that are absent from the training dataset.

	Across the six hold-out compositions the predicted curve $\hat y(t)$ coincides with the simulated transmission throughout the thicknesses of interest. The two slightly harder cases, Bi$_{2}$O$_{3}$-BaSO$_{4}$ and Bi-W (both at \SI{80}{\kilo\volt}), track the simulated curve in shape with only a small offset, while the four remaining cases are essentially indistinguishable from the simulation.

    The held-out compositions above still draw every constituent from the twelve materials seen during training, only their \emph{weight ratio combination} is new. We wanted to test STAR-XPC not only with the similar materials but also with unseen materials. To probe the molecular embedding pipeline itself and the generalization to unseen materials, we ran  additional PENELOPE/penEasy simulations for lead (Pb) and gadolinium (Gd), which were absent (unseen) in the training phase. For these runs we considered a single shielding material with $70$~wt\% in polyethylene (the discrete $r_{\mathrm{poly}}=0.30$) at \SI{100}{\kilo\volt}, swept over ten thicknesses from $0.15$ to $1.5$~mm. All other inputs (tube voltage, thickness range, mass-fraction split and filler cardinality) stay within the trained distribution, isolating the test to the molecular identity of Pb and Gd. Their embeddings are computed with the identical ChemBERTa pipeline from their canonical bracketed SMILES (\texttt{[Pb]}, \texttt{[Gd]}), then passed through the STAR-XPC with no retraining or fine-tuning.
    %%%%%%%%%%%%%%%%%%%%%%%%%%%%%%%%%%%%%%%%%%%%
    \begin{figure}[pos=!htb]
		\centering
		\begin{subfigure}{0.49\linewidth}
			\includegraphics[width=\linewidth]{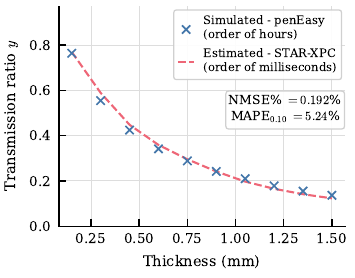}
			\caption{Pb; $70/30$~wt\% Pb/(C$_2$H$_4)_n$}
		\end{subfigure}\hfill
		\begin{subfigure}{0.49\linewidth}
			\includegraphics[width=\linewidth]{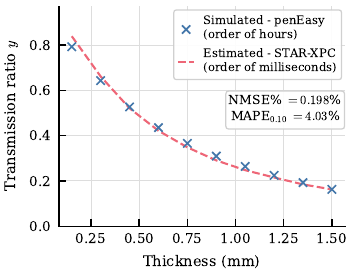}
			\caption{Gd; $70/30$~wt\% Gd/(C$_2$H$_4)_n$}
		\end{subfigure}
		\caption{Transmission ratio versus thickness at \SI{100}{\kilo\volt} for two shielding elements entirely absent from the training vocabulary.}
		\label{fig:pbgd_curves}
	\end{figure}

	Figure~\ref{fig:pbgd_curves} shows both thickness sweeps. Despite zero exposure to Pb or Gd during training, STAR-XPC tracks both curves closely: $\mathrm{NMSE\%}=0.192\%$ ($\mathrm{MAPE}_{0.10}=5.24\%$) for Pb and $\mathrm{NMSE\%}=0.198\%$ ($\mathrm{MAPE}_{0.10}=4.03\%$) for Gd. This is a  harder generalization task compared to previous held-out compositions of similar materials. The model has no prior exposure to either element's embedding at all, whereas the held-out compositions test still reuses embeddings seen during training. On these two materials, Pb and Gd, this is an evidence against a pure lookup-table interpretation of the embedding pipeline.

    %%%%%%%%%%%%%%%%%%%%%%%%%%%%%%%%%%%%%%%%%%%%%%%%%%%%%%%%%%%
    %%%%%%%%%%%%%%%%%%%%%%%%%%%%%%%%%%%%%%%%%%%%%%%%%%%%%%%%%%%
    %%%%%%%%%%%%%%%%%%%%%%%%%%%%%%%%%%%%%%%%%%%%%%%%%%%%%%%%%%%
	\subsection{Discussion}\label{sec:discussion}

    \subsubsection{Set Transformer Architecture Benefits}

    We chose the set transformer architecture due to benefits like permutation invariance, self-attention capabilities, and linear scaling via inducing points.

    The bulk attenuation of a homogenized mixture is independent of the lexicographic order of its constituents. The Set Transformer enforces this symmetry by construction and this is consistent with the observation that the model handles $1$-, $2$-, and $3$-component mixtures with the same network. The training data contain at most three shielding materials per sample. The input format admits up to ten shielding constituents plus the polymer, so mixtures with more than three materials can be passed to the same network without modification; however, such mixtures would also contain fewer padding tokens than any training sample, and since the unmasked padding tokens carry the global descriptors and also receive attention, accuracy on higher-order mixtures cannot be assumed and remains to be validated. Likewise, the trained network is tied to the set size $N=11$; changing $N$ changes the number of padding tokens and therefore the prediction. The ISAB mechanism keeps the attention cost linear in the set size, so a variant with masked padding and a dedicated global-descriptor token could be scaled to larger mixtures, which we leave to future work.

	\subsubsection{Limitations and Challenges}

	Because the polymer is identical across all samples (polyethylene, (C$_{2}$H$_{4})_n$, encoded in the embedding pipeline by the SMILES \texttt{C=C}), the polymer token's embedding has zero discriminative power. The bulk effect of the polymer enters exclusively through the four global descriptors and through its mass fraction. A study that varies the polymer would require a richer embedding pipeline, e.g.\ a polymer-aware encoder fine-tuned on engineering polymers, and is beyond the scope of this paper.

    A related concern is the domain mismatch of the molecular embedding itself. ChemBERTa-zinc-base-v1~\citep{chithrananda2020chemberta} is pretrained on ZINC15~\citep{sterling2015zinc15}, a corpus of drug-like organic molecules. Several of our shielding constituents (e.g.\ elemental \texttt{[Bi]}, \texttt{[W]}, or inorganic oxides such as Bi$_{2}$O$_{3}$) are far outside that pretraining distribution, and the resulting embeddings are not validated to carry chemically meaningful structure for these species. With only $13$ distinct constituent molecules in the dataset (Section~\ref{sec:baselines}), the embedding pipeline could in principle function close to a per-molecule lookup table rather than exploiting ChemBERTa's generalization. Section~\ref{sec:pbgd} provides an evidence against a pure lookup-table interpretation, since the model generalizes to two elements (Pb, Gd) never seen during training.

	\section{Conclusions and Future Directions}\label{sec:conclusions}

	We have presented a permutation-invariant Set Transformer model for X-ray energy transmission in polyethylene/shielder composites, trained on a PENELOPE/penEasy-generated dataset of $n=4697$ simulated configurations spanning four tube voltages and $12$ main shielding material chemistries. Cross-validated over all four voltages, the model achieves $\mathrm{NMSE\%}=0.0294\%$ ($R^{2}=0.9988$) with $\mathrm{NMSE\%} \le 0.129\%$ on six held-out compositions never seen during training and $\mathrm{NMSE\%} \le 0.2\%$ on two unseen materials (Pb and Gd) to show its generalization capabilities. We have open data and open codes on GitHub (\url{https://github.com/cmpe-netlab41/attenupy}) and HuggingFace (\url{https://huggingface.co/cmpe-netlab41/STAR-XPC}), so that researchers focusing on estimating X-Ray Transmission may use STAR-XPC easily with example codes.

	Each architectural choice corresponds to a physical property of the problem. Permutation invariance reflects the fact that the order of the constituents does not change the attenuation of the material. The attention-based ISAB encoder lets the constituents influence one another, capturing interactions that a simple linear mixture rule cannot. With these architectural choices, the proposed model combines accuracy comparable to Monte Carlo simulations with an inference cost orders of magnitude lower (sub-millisecond per query on a single GPU vs.\ hours per Monte Carlo run), opening the door to interactive composite design for medical imaging, industrial radiography, and shielding optimization. In the proposed model, only the transmission ratio is predicted. For future directions, we plan to focus on other tally metrics that can also be added to the output head of the ML-surrogate. We also plan to investigate physics-informed regularizers~\citep{raissi2019pinns}.

    \section*{Generative AI Usage Statement}
    Generative AI tools were used for coding assistance and language polishing. All contents have been carefully reviewed and verified by the authors.

\bibliographystyle{cas-model2-names}

\bibliography{myrefs}

\end{document}